\documentclass[conference]{IEEEtran}
\IEEEoverridecommandlockouts
\usepackage{cite}
\usepackage{amsmath,amssymb,amsfonts}
\usepackage{algorithmic}
\usepackage{graphicx}
\usepackage{textcomp}
\usepackage{balance}
\usepackage{color}
\usepackage{booktabs}

\def\BibTeX{{\rm B\kern-.05em{\sc i\kern-.025em b}\kern-.08em T\kern-.1667em\lower.7ex\hbox{E}\kern-.125emX}}

\begin{document}

\title{Analysis of adversarial attacks against \\ CNN-based image forgery detectors\\
{\footnotesize
}
\thanks{This material is based on research
sponsored by the Air Force Research Laboratory and the Defense Advanced Research Projects Agency under agreement number FA8750-16-2-0204.
The U.S. Government is authorized to reproduce and distribute reprints for Governmental purposes notwithstanding any copyright notation thereon.
The views and conclusions contained herein are those of the authors
and should not be interpreted as necessarily representing the official policies or endorsements, either expressed or implied,
of the Air Force Research Laboratory and the Defense Advanced Research Projects Agency or the U.S. Government.}
}

\author{\IEEEauthorblockN{Diego Gragnaniello, Francesco Marra, Giovanni Poggi, Luisa Verdoliva}
\IEEEauthorblockA{\textit{Department of Electrical Engineering and Information Technology} \\
\textit{University Federico II of Naples, Naples, Italy} \\
name.surname@unina.it}
}

\maketitle

\begin{abstract}
With the ubiquitous diffusion of social networks, images are becoming a dominant and powerful communication channel.
Not surprisingly, they are also increasingly subject to manipulations aimed at distorting information and spreading fake news.
In recent years, the scientific community has devoted major efforts to contrast this menace, and many image forgery detectors have been proposed.
Currently, due to the success of deep learning in many multimedia processing tasks,
there is high interest towards CNN-based detectors, and early results are already very promising.
Recent studies in computer vision, however, have shown CNNs to be highly vulnerable to adversarial attacks,
small perturbations of the input data which drive the network towards erroneous classification.
In this paper we analyze the vulnerability of CNN-based image forensics methods to adversarial attacks,
considering several detectors and several types of attack,
and testing performance on a wide range of common manipulations, both easily and hardly detectable.
\end{abstract}


\section{Introduction}

In the era of social networks, images have become a dominant communication vehicle.
They convey information with higher immediacy and depth than text, and have the potential to elicit strong responses in the observers.
Unfortunately,
with modern media editing tools, tampering with images has become very easy.
The manipulated images can be used to discredit people, direct public opinion, even change the course of political events,
and pass easily unnoticed from ordinary people.

A number of multimedia forensic tools have been proposed in the last years to detect image manipulations \cite{Korus2017}.
In particular, methods based on high-order statistics of image residuals have drawn great attention since long time \cite{Farid2003, Bayram2006}.
Indeed, when a pristine image is modified, by inserting or removing objects, or modifying global characteristics,
several low-level operations are usually involved, like linear or non-linear filtering, resizing, or compression.
All these operations leave subtle but distinctive traces in the image micro-patterns,
which can be discovered by means of suitable image descriptors extracted from the high-pass image residual.
To this end, the SPAM (subtractive pixel adjacency matrix) features \cite{Pevny2010}
and the SRM (spatial rich models) \cite{Fridrich2012}
have shown great potential for many image forensics tasks \cite{Cozzolino2014a, Boroumand2017, Li2018}.
In particular,  excellent results \cite{Cozzolino2014a, Cozzolino2015b, Li2018}
can be obtained even by considering one specific single model from \cite{Fridrich2012},
the one computing 4-pixel co-occurrences on the residuals of 3rd order linear filter.
Given their similarity wirh SPAM features,
here for the sake of brevity we will refer to them as S3SPAM or simply SPAM features.

Nonetheless, the current trend in forensics, and in multimedia processing in general, is to abandon handcrafted features in favor of deep learning.
Given a sufficiently large training set,
deep nets, typically convolutional neural networks (CNN), learn from the data which features best address the given task,
reaching usually impressive performance.
The first CNN-based detector of image manipulation was proposed in \cite{Bayar2016},
inspired to previous work in steganalysis.
Its main peculiarity is an {\it ad hoc} first layer, comprising a bank of filters constrained to extract high-pass features.
Since the most relevant information for discrimination is hidden in the high-pass image content,
such filters speed up convergence to a satisfactory solution.
In \cite{Cozzolino2017}, instead, it was proven that S3SPAM features can be extracted by a simple shallow CNN.
Besides reproducing the very good results of the original detector,
the resulting net can be further improved by fine tuning on a specific dataset, providing a very good performance even with a small training set.
Very recently,
another deep learning solution has been proposed, aimed at detecting the processing history of JPEG images \cite{Boroumand2018}.

All the above networks, though very effective, are relatively shallow.
Very deep architectures can be expected to provide a further performance boost.
Tellingly,
in a recent competition on camera model identification
organized by the IEEE Signal Processing society on the Kaggle platform\footnote{https://www.kaggle.com/c/sp-society-camera-model-identification},
all top-ranking teams proposed solutions based on an ensemble of very deep networks.
Likewise, very deep networks have shown top performance and higher robustness \cite{Marra2018}
in detecting images manipulated by generative adversarial networks (GAN).

\begin{figure*}[t!]
	\centering
	\includegraphics[width=0.9\linewidth]{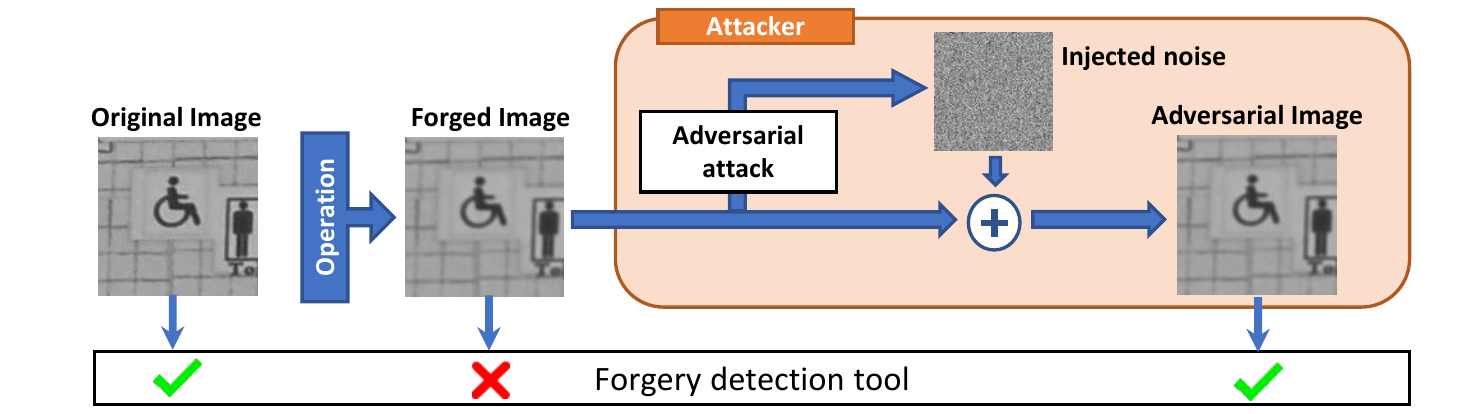}
	\caption{Our reference scenario. The subtle traces left during image forgery can be detected by a forensic tool (red cross in the bottom box).
    However, an attacker can conceal such traces by injecting suitable adversarial noise, thus misleading the detector into authenticating the image as pristine (green checkmark).}
	\label{fig:scenario}
\end{figure*}

Although deep learning holds great potential for multimedia forensics, one should not rely on a safe environment, counting on the attacker's naivete.
On the contrary, the risks incurred by counter-forensic actions, aimed at neutralizing forensic tools (see Fig.1),
must be taken into serious account and analyzed in depth.
Some recent papers \cite{Marra2015,Chen2017}, for example, propose to attack SPAM-based detectors
by means of iterative gradient descent algorithms, which prove very effective, although definitely slow.
Attacking CNNs, however, has proven to be much simpler \cite{Goodfellow2015}.
By exploiting the intrinsic differentiability of the loss function,
a suitable adversarial noise can be easily generated and added to the input image to modify the network decision, without visible image impairments.
Following this seminal paper, many more attacks based on adversarial noise have been devised.
In addition, deep learning can be used itself for counter-forensics.
In \cite{Kim2018}, a GAN-based architecture was proposed to conceal the traces of 3$\times$3 median filtering.
Such a study, though limited to a very special case, opens the way to interesting developments.

Here, we investigate on the effectiveness of adversarial attacks to CNN-based detectors.
We consider a large set of manipulations, both easily detectable and more challenging, and several CNN-based detectors.
Specific adversarial noise is generated for each detector, and the effects are assessed both on the target detector and on non-target ones.
The performance of GAN-based restoration is also assessed, with reference to the especially challenging case of median filtering.
To the best of our knowledge, this is the first study on this topic.

In the rest of the paper,
we describe the detectors (Section 2), the attacks (Section 3), and the experimental analysis (Section 4), before drawing conclusions (Section 5).

\section{CNN-based detectors of image manipulation}

In this Section we briefly recall some relevant CNN-based detectors, with their main features.
However we also consider a baseline conventional detector, using handcrafted features \cite{Fridrich2012} and support vector machine (SVM) classification.

\subsection{SPAM+SVM}
To extract the residual features proposed in \cite{Fridrich2012}
the original image is high-pass filtered and quantized with a small number of bins.
Then, co-occurrences are computed, encoded, and collected in a linear histogram feature, normalized to unit energy.
Depending on the specific parameters of this process, different features are obtained,
collectively called rich models \cite{Fridrich2012}.
As said before,
we consider one single  model here, with third-order linear filter, 5-bin quantization, and 4-lag co-occurrences
({\em s3\_spam14hv}) and will refer to it as SPAM features from now on.

\subsection{Bayar2016}
In \cite{Bayar2016} a relatively small CNN is proposed for image manipulation detection, referred to as Bayar2016 from now on,
comprising three convolutional layers, two max-pooling layers, and three fully-connected layers.
In order to immediately extract residual-based features, as suggested by the literature,
filters of the first layer, with 5$\times$5 receptive field, are constrained to respect the following rule
\begin{equation}
\left\{ \begin{array}{l}
    w(0,0) = -1 \\
    \sum_{l,m \neq 0} w(l,m) = 1
    \end{array} \right.
\end{equation}
Therefore, the sum of all weights is 0, enforcing the high-pass nature of the filters.
In particular, the off-center pixels are combined to compute a prediction of the center pixel, so the output of the filter can be regarded as a prediction error.

\subsection{Cozzolino2017}
The main result of \cite{Cozzolino2017} is that a large class of conventional features can be computed exactly by suitable convolutional networks.
Although the result is quite general, the work focuses on the SPAM feature described before.
Exact SPAM feature extraction requires only two convolutional layers, followed by hardmax and average pooling.
The extracted features could then be used to train an external SVM.
However, a full-fledged CNN-based detector is also built in \cite{Cozzolino2017},
by complementing the feature extractor subnet with a fully connected layer which replaces the external SVM classifier.
Then, the hardmax is also replaced by softmax to ensure differentiability, allowing further training by backpropagation.
Besides the theoretical result, the CNN proposed in \cite{Cozzolino2017} can faithfully replicate the SPAM-SVM suite,
and improve upon it by means of quick fine tuning over a very small training set.
This latter version, referred to as Cozzolino2017, is considered here.

\subsection{Very deep nets: Xception}

In recent years, a large experimental evidence has accumulated showing that network depth plays a fundamental role for generalization ability.
State-of-the-art architectures in computer vision and related fields,
such as ResNet, DenseNet, InceptionNet, XceptionNet, all comprise from several dozens to hundreds of layers.
Our own experience in forensic applications \cite{Marra2018} confirms the superior robustness of deep nets to challenging and off-training conditions.
On the down side, deep nets require very large datasets for correct training, a condition not always met in practice.

To include a deeper net in our comparative assessment we selected Xception \cite{Chollet2017},
comprising a total of 42 layers, 36 convolutional, 5 pooling, and one fully connected.
Its main architectural innovation is the use of separable filters.
That is, 3D convolutions are obtained by the cascade of 2D spatial and 1D cross-map convolutions.
Thanks to this constraint,
the number of free parameters drops significantly w.r.t. competing nets or, under a different point of view,
a deeper architecture can be adopted for the same level of complexity,
allowing the use of such a deep net even with a relatively small training set.

\section{Attacking forensic detectors}

In this Section, we describe some possible strategies to attack image forensic detectors, in particular
\begin{itemize}
\item   gradient descent algorithms for SPAM+SVM;
\item   generation of adversarial noise for CNN-based detectors;
\item   GAN-based restoration of manipulated images.
\end{itemize}
Although we focus on targeted attacks, designed against a specific detector,
universal counter-forensic methods are also studied, {\it e.g.}, \cite{Barni2013b}

\subsection{Attacking a SPAM-based detector by gradient descent}

\begin{figure}[t!]
	\centering
	\includegraphics[width=1\linewidth]{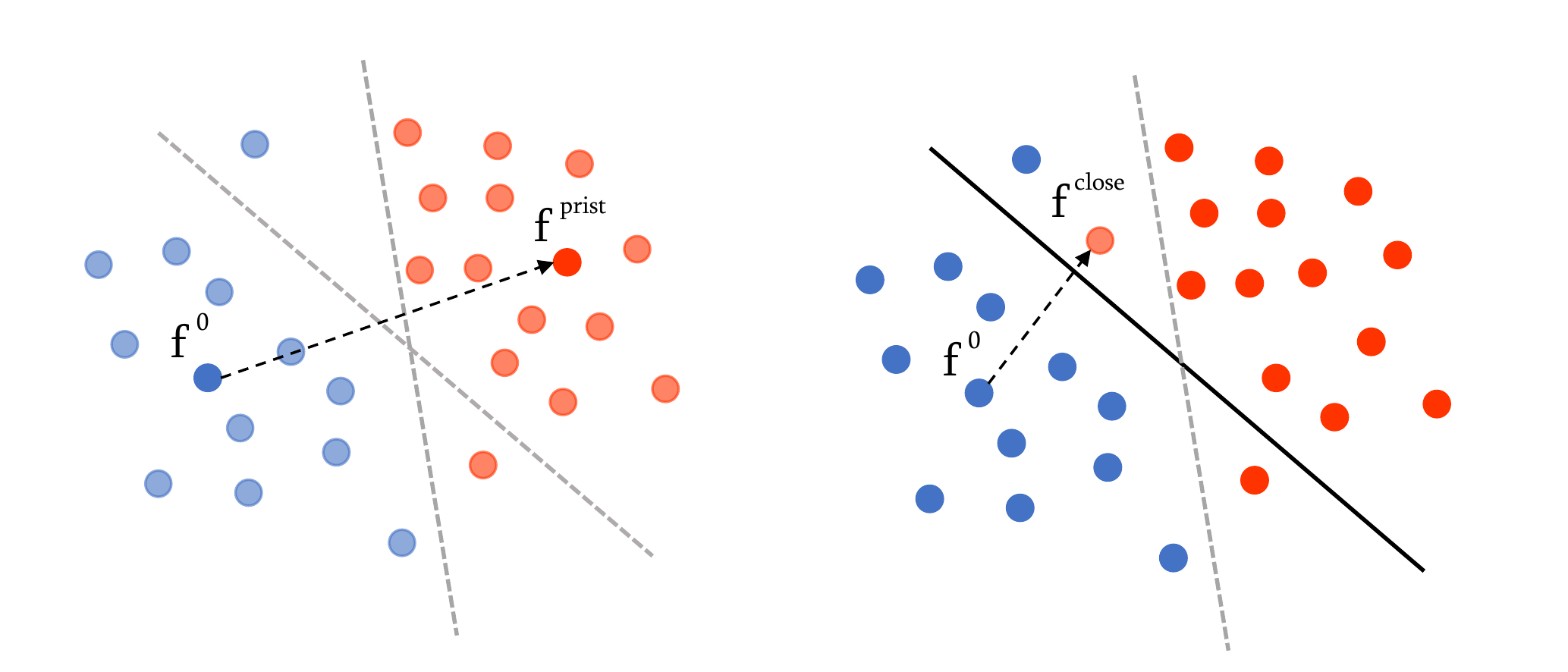}
	\caption{Attacks in the feature space. Left: restoring the feature of the pristine image. Right: crossing the decision boundary of the target detector.
    The second attack is simpler but may fail with non-target detectors.}
	\label{fig:Marra_attacks}
\end{figure}

\newcommand{\prist}{{prist}}
Let $X^\prist$ and $X^0$ be the pristine and manipulated images, with $f^\prist$ and $f^0$ the corresponding feature vectors, SPAM in our case.
Lacking perfect knowledge on the classifier,
the attacker wants to modify $X^0$ into a new image, $\hat{X}$, similar to $X^0$, to limit distortion,
but whose feature, $\hat{f}$, is so close to $f^\prist$ to fool the detector, see Fig.\ref{fig:Marra_attacks}(left).
In formulas, the problem can be cast as
\begin{equation}
    \hat{X} = \arg\min_X \psi(f(X),f^\prist), \hspace{4mm} {\rm s.t.}\;\; \phi(X,X^0)<T
    \label{eq:formulation3}
\end{equation}
where $\phi(\cdot,\cdot)$ and $\psi(\cdot,\cdot)$ are image and feature space distances, and $T$ a suitable threshold on distortion.

In \cite{Marra2015} an iterative algorithm is proposed,
where the objective function is minimized through local changes on the image, like in the iterated conditional modes method.
This approach is effective but quite slow,
because the feature must be recomputed at each new step, due to its complex nonlinear relationship with the image.

Note that, if the classifier is perfectly known,
one can target $f^{close}$, the feature closest to $f^0$ across the decision boundary, rather than $f^\prist$, as shown in Fig.\ref{fig:Marra_attacks}(right).
This speeds up convergence considerably, but reduces robustness with respect to off-target detectors, as also depicted in Fig.\ref{fig:Marra_attacks}(right).

\subsection{Attacking CNNs by adversarial noise}

\begin{table*}[ht!]
\centering
\caption{Performance of several detectors in the presence of common manipulations. Top: easy cases. Bottom: challenging cases.}
{\small
\begin{tabular}{l|rrr|rrr|rrr|rrr} \toprule
                             & \multicolumn{3}{c|}{SPAM} & \multicolumn{3}{c|}{Bayar2016} & \multicolumn{3}{c|}{Cozzolino2017} & \multicolumn{3}{c}{Xception} \\
Manipulation/parameter       &   FPR &    TPR &   ACC &   FPR &   TPR &   ACC &  FPR &    TPR &   ACC &   FPR &   TPR &   ACC \\ \midrule
Blurring, $\sigma$=1.10      &  0.06 & 100.00 & 99.97 &  0.02 & 99.98 & 99.98 & 0.07 & 100.00 & 99.96 &  2.26 & 99.87 & 98.81 \\
JPEG compression, $Q$=70     &  0.02 & 100.00 & 99.99 &  0.83 & 99.69 & 99.43 & 0.00 &  99.98 & 99.99 &  0.63 & 98.54 & 98.95 \\
Median filtering, 7$\times$7 &  0.54 &  99.90 & 99.68 &  0.56 & 99.93 & 99.69 & 1.26 & 100.00 & 99.37 &  0.85 & 99.96 & 99.56 \\
Resizing, scale=1.500        &  0.26 & 100.00 & 99.87 &  0.50 & 99.94 & 99.72 & 0.00 &  99.96 & 99.98 &  6.56 & 99.11 & 96.28 \\ \midrule
Blurring, $\sigma$=0.50      & 24.02 &  99.12 & 87.55 &  8.74 & 97.93 & 94.59 & 7.69 &  99.06 & 95.69 & 13.76 & 88.06 & 87.15 \\
JPEG compression, $Q$=90     &  6.46 &  88.26 & 90.90 &  0.72 & 90.17 & 79.72 & 5.83 &  94.81 & 94.49 &  2.93 & 90.37 & 93.72 \\
Median filtering, 3$\times$3 &  0.06 &  99.80 & 99.87 &  0.31 & 99.91 & 99.80 & 0.43 &  99.91 & 99.74 &  7.46 & 99.54 & 96.04 \\
Resizing, scale=1.010        &  9.07 &  99.29 & 95.11 &  2.72 & 99.59 & 98.44 & 3.00 &  99.67 & 98.33 &  8.26 & 98.22 & 94.98 \\ \bottomrule
\end{tabular}
}
\label{tab:no_attack}
\end{table*}

Experiments in computer vision \cite{Goodfellow2015,Papernot2016JSMA}
have clearly established the vulnerability of CNN-based detectors to adversarial attacks.
A suitable adversarial noise pattern can be added to the input image to mislead the classifier, see Fig.1, without impairing its visual quality.
With CNNs, some simple algorithms for the generation of adversarial noise are available.

The Fast Gradient Sign Method (FGSM), proposed in \cite{Goodfellow2015},
exploits the differentiability of the loss function.
The gradient of the loss with respect to each pixel of the input image is first computed by backpropagation.
Then, each pixel is modified by a small quantity, $\pm\epsilon$, taking the sign of the local gradient.
Neglecting higher order effects, all perturbations increase the loss,
and hence a large change in output can be obtained with very low-variance adversarial noise.

Following this early, and simple, method, more sophisticated solutions have been proposed.
DeepFool \cite{Moosavi2016} is based on a local linearization of the classifier under attack,
which allows one to project the input image on the approximate decision boundary, and to introduce the minimum perturbation necessary to cross it.
The Jacobian-based Saliency Map Attack (JSMA) \cite{Papernot2016JSMA} relies on a greedy iterative procedure.
Unlike FGSM, it attacks only the pixels that contribute most to the correct classification,
identified by a suitable saliency map.
In \cite{Madry2017}, adversarial noise generation is formulated as a min-max optimization,
with the double aim of generating effective adversarial examples and training robust classifiers.
The resulting algorithm, projected gradient descent (PGD), provides the optimum adversarial examples when the network is perfectly known.
Noteworthy, FGSM can be regarded as a single-step scheme to solve the maximization step of PGD.

In the experiments, we will consider only the FGSM algorithm,
because of its low complexity (JSMA and PGD are orders of magnitude slower) and easy interpretation.
Note that, in a realistic setting, images must be rounded to integer values to be stored or transmitted,
so, unlike in theoretical analyses, we consider only integer values for the $\epsilon$ parameter.

\begin{table}[t!]
	\centering
	\caption{TPR with adversarial noise (FGSM, $\epsilon=1$). Target: Bayar2016.}
	{\small
		\begin{tabular}{l|r|r|r|r} \toprule
			Manipulation      &      SPAM & Bayar2016 & Cozz.2017 & Xception \\ \midrule
			Blurring 1.10     &      0.19 & {\bf 0.00}&      0.00 &     3.51 \\
			JPEG 70           &      0.06 & {\bf 0.00}&      0.08 &    10.22 \\
			Median 7$\times$7 &     99.16 & {\bf 0.00}&     99.96 &    26.56 \\
			Resizing 1.50     &      0.02 & {\bf 0.00}&      0.00 &     3.28 \\ \midrule
			Blurring 0.50     &      1.43 & {\bf 0.00}&      7.67 &    12.17 \\
			JPEG 90           &      0.26 & {\bf 0.02}&      0.13 &    18.01 \\
			Median 3$\times$3 &     86.65 & {\bf 0.00}&     45.43 &    10.49 \\
			Resizing 1.01     &      0.95 & {\bf 0.00}&      5.92 &    10.18 \\ \bottomrule
		\end{tabular}
	}
	\label{tab:target_Bayar}
\end{table}

\begin{table}[t!]
	\centering
	\caption{TPR with adversarial noise (FGSM, $\epsilon=1$). Target: Cozz.2017.}
	{\small
		\begin{tabular}{l|r|r|r|r} \toprule
			Manipulation      &      SPAM & Bayar2016 &  Cozz.2017 & Xception \\ \midrule
			Blurring 1.10     &     32.63 &     32.68 & {\bf 32.19}&    32.75 \\
			JPEG 70           &      0.20 &     13.51 & {\bf  0.00}&    17.16 \\
			Median 7$\times$7 &     98.78 &     93.34 & {\bf 88.20}&    96.71 \\
			Resizing 1.50     &     18.76 &     18.92 & {\bf 18.91}&    18.58 \\ \midrule
			Blurring 0.50     &      1.05 &      1.66 & {\bf 15.67}&     4.90 \\
			JPEG 90           &      0.02 &     27.94 & {\bf  0.00}&    16.42 \\
			Median 3$\times$3 &     92.29 &     99.95 & {\bf  6.43}&    35.22 \\
			Resizing 1.01     &      0.71 &      4.49 & {\bf 13.89}&     9.02 \\ \bottomrule
		\end{tabular}
	}
	\label{tab:target_Cozzolino}
\end{table}

\begin{table}[h!]
	\centering
	\caption{TPR with adversarial noise (FGSM, $\epsilon=1$). Target: Xception.}
	{\small
		\begin{tabular}{l|r|r|r|r} \toprule
			Manipulation      &      SPAM & Bayar2016 & Cozz.2017 &   Xception \\ \midrule
			Blurring 1.10     &      8.33 &     12.50 &      0.63 & {\bf  0.00}\\
			JPEG 70           &      3.00 &     38.85 &      0.81 & {\bf  0.41}\\
			Median 7$\times$7 &     99.30 &     24.74 &    100.00 & {\bf  0.00}\\
			Resizing 1.50     &     11.56 &     13.22 &      1.06 & {\bf  0.00}\\ \midrule
			Blurring 0.50     &     26.31 &     30.41 &     17.20 & {\bf  0.00}\\
			JPEG 90           &      2.24 &     33.46 &      0.33 & {\bf 17.98}\\
			Median 3$\times$3 &     99.83 &     43.93 &    100.00 & {\bf  7.50}\\
			Resizing 1.01     &     14.24 &     27.91 &      8.70 & {\bf  0.00}\\ \bottomrule
		\end{tabular}
	}
	\label{tab:target_Xception}
\end{table}

\subsection{Attacks based on GANs}

Starting from the 2015 seminal work of Goodfellow {\it et al.} \cite{Goodfellow2015}
generative adversarial networks have gained a major role in deep learning,
providing remarkable results in a large number of tasks involving image synthesis and/or manipulation.
The basic idea is to train in parallel two competing nets,
a generator, which tries to synthesize images with a natural appearance, and a discriminator, which tries to tell apart natural from synthetic images.
This competition gradually improves the performance of both nets.
Ideally, at convergence, the generator should be able to produce images that are indistinguishable from natural ones.

Recently, a GAN-based method has been proposed \cite{Kim2018} for the restoration of median filtered images.
For this application, the generator does not start from a random noise vector to synthesize the output,
as usual with GANs, but takes in input the manipulated image and restores its natural features.
Accordingly, the generator loss includes not only an adversarial term, which measures its ability to fool the detector, but also two image quality terms.
These measure objective quality (distance from the original) and perceptual quality of the generated image.
We refer the reader to the original paper for all details of the method,
underlining only that the generator relies heavily on residual connections to improve stability and speed up convergence.

\section{Experimental analysis}

To carry out our experimental analysis we generate a dataset
taking 200 images from each of 9 different devices, and 192 partially overlapping $128\times 128$ patches from each image.
We consider 4 types of image manipulation: Gaussian blurring, JPEG compression, median filtering, and resizing,
with two different settings for each case corresponding to ``easy'' and ``challenging'' tasks.
For example, a Gaussian filter with $\sigma=1.10$ causes easily detectable blurring, unlike with $\sigma=0.50$.
In each binary classification task,
patches from 6 devices chosen at random are used for training, the others for testing.
Overall, each training set comprises more than 200k+200k patches, still relatively small for deep learning applications.

In Tab.\ref{tab:no_attack} we report, for the considered detectors,
false positive rate (FPR), true positive rate (TPR), and overall accuracy (ACC), in the absence of counter-forensic attacks.
For easy cases (top), accuracies are always close to 100\%,
only Xception shows a somewhat worse performance, very likely due to the limited training set.
For more challenging manipulations (bottom), larger differences are observed,
with some poorer results on JPEG compression (Bayar2016, SPAM) and Gaussian blurring (SPAM, Xception).
Nonetheless, a very good detection performance is still observed, in general.

In Tables {\ref{tab:target_Bayar} through \ref{tab:target_Xception}} we study the case in which adversarial noise is added to the manipulated images, using FGSM with $\epsilon=1$, namely, the weakest adversarial noise which survives the image rounding.
Since neither the pristine images nor the detectors change, we report only the TPR for the attacked manipulated images.
Note also that the PSNR is always 48.13 dB (MSE=1), hence no visual impairment can be appreciated.
The attack is very effective when the same net is used to generate the adversarial noise and to detect the manipulation (boldface entries).
Only Cozzolino2017, and only for the 7$\times$7 median filtering, keeps providing a good TPR.
In the absence of alignment, however, the attack is much less effective, especially for median filtering, both 3$\times$3 and 7$\times$7,
for which both SPAM and Cozzolino2017 provide a TPR close to 100\%.
These results suggest that, at least in such cases,
the adversarial noise is not restoring the features of pristine images disrupted by the manipulation, but only exploiting some detector weaknesses.

This latter consideration further motivates us to explore the GAN-based attack, which has the very goal of restoring manipulated images.
In Tab.\ref{tab:GAN_attack} we report results only for the critical median filtering cases.
They seem to confirm a better ability of the GAN-based method to attack uniformly all detectors.
Actually, the original architecture proposed in \cite{Kim2018} works well only in the 3$\times$3 case, and never fools Xception.
However, if we replace the original discriminator with a VGG net \cite{Simonyan2014},
the attack becomes more effective for all detectors, and none of them reaches a 50\% TPR.

\begin{table}[]
\centering
\caption{TPR for median filtering after GAN-based restoration. Top: Kim2018 discriminator. Bottom VGG discriminator.}
{\small
\begin{tabular}{l|r|r|r|r} \toprule
Manipulation       &      SPAM & Bayar2016 & Cozz.2017 &   Xception  \\ \midrule
M-7$\times$7 (Kim) &     70.44 &     97.78 &     89.39 &      89.19  \\
M-3$\times$3 (Kim) &     19.30 &     49.20 &     15.69 &      91.70  \\ \midrule
M-7$\times$7 (VGG) &      3.74 &     49.15 &     18.48 &      35.56  \\
M-3$\times$3 (VGG) &      1.17 &      0.74 &      0.78 &      23.76  \\ \bottomrule
\end{tabular}
}
\label{tab:GAN_attack}
\end{table}

\section{Conclusions}
We have presented an investigation on adversarial attacks to CNN-based image manipulation detectors.
Even a rather simple attack can completely mislead the target detector and largely reduce the detection performance of off-target detectors.
As only exception, the adversarial noise attack was not able to conceal 7$\times$7 median filtering, which deeply modifies the image fine structures.
However, a suitable GAN-based attack proves to work well even in this challenging case.

Obviously, these early results represent only a proof of concept,
and more thorough analyses are necessary to gather a solid understanding of the relevant issues.
More sophisticated attacks must be considered, and more detectors tested, on a wider range of manipulations.
In particular, realistic applications over social networks, involving resizing and compression, should be considered.

\balance
\bibliographystyle{IEEEtran}
\bibliography{refs}

\end{document}